\documentclass[runningheads]{llncs}
\usepackage{graphicx}
\usepackage{amsmath}
\usepackage{booktabs}
\usepackage{multirow}
\usepackage{subfigure}
\usepackage{cite}
\begin{document}
\title{An Improvement of PAA on Trend-based Approximation for Time Series}
%
%
\author{Chunkai Zhang\inst{1} \and
Yingyang Chen\inst{2} \and
Ao Yin\inst{3} \and
Zhen Qin\inst{4} \and
Xing Zhang\inst{5} \and
Keli Zhang\inst{6}\and
Zoe L. Jiang\inst{7}
}
\authorrunning{Yingyang Chen et al.}
%
\institute{ Department of Computer Science and Technology, \\
	Harbin Institute of Technology, Shenzhen, China\inst{1,2,3,4,7} \\
	Engineering Laboratory for Big Data Collaborative Security Technology 
	\inst{5,6}
\email{ckzhang812@gmail.com\inst{1}, yingyang\_chen@163.com\inst{2}, 
yinaoyn@126.com\inst{3}, 
qinzhen\_qd@163.com\inst{4},zhangxing@cecgw.cn\inst{5},zhangkeli@cecgw.cn\inst{6},
  \\
zoeljiang@hit.edu.cn\inst{7}}
}
\maketitle              
\begin{abstract}
Piecewise Aggregate Approximation (PAA) is a competitive basic dimension 
reduction method for high-dimensional time series mining. When deployed, 
however, the limitations are 
obvious that 
some important information will be missed, especially the trend.
In this paper, we propose two new approaches for time series that utilize 
approximate trend feature information. Our first method is based 
on relative mean value 
of each segment to
record the trend, which divide each segment into 
two parts and use the numerical average respectively to represent 
the trend. We proved that this method 
satisfies lower bound which guarantee no false dismissals.
Our second method uses a binary string to record the trend which is also 
relative to mean in each segment. 
Our methods are applied on similarity measurement in classification and anomaly 
detection, 
the experimental results show the improvement of accuracy and effectiveness by 
extracting the trend feature suitably.

\end{abstract}

\section{Introduction}

Time series is a series of data points indexed in time order, 
which is widely existed in fields of 
medical\cite{Hu2016The,Dersch2002Cluster,Zhang2017Design}, 
business\cite{Rui2012A}, industry\cite{Storer2017LittleTable,Yu2016An}, 
cyber security\cite{Yong2007A,Rodriguez2010Improving} and so on. Time series 
mining is one of the attractive research topics and a key 
issue for the last decade, such as 
classification\cite{Xi2006Fast}, 
clustering\cite{Paparrizos2016k}, anomaly 
detection\cite{Shokoohi2015Discovery,trident}, time series 
visualization\cite{Himberg2004Validating,Landesberger2016MobilityGraphs}. 
Most of mining technologies require comparison of similarity measurement, which 
can effect the accuracy and efficiency of mining. A series of measurements 
have been proposed, such as Manhattan Distance\cite{Yi2000Fast}, Euclidean 
Distance\cite{faloutsos1994fast}, Chebyshev Distance\cite{Cantrell2001Modern}.
The typical measure is Euclidean Distance(ED), which is the sum of straight 
line distance between two points through time series. 

Time series is a high-dimensional data that leads to expensive time and space 
cost when processed with the raw data directly by using ED, so dimensionality 
reduction is required to improve the efficiency.
There has been much work in dimensional reduction, and one of the popular 
approaches is using spatial method to index the data in the transformed space 
including Discrete Fourier Transform (DFT)\cite{faloutsos1994fast,Lin2003A}, 
Singular Value Decomposition 
(SVD)\cite{faloutsos1994fast,Keogh2001Dimensionality}, Discrete Wavelet 
Transform (DWT)\cite{Chan1999Efficient,Kahveci2002Variable}. And 
there are piecewise aggregate representation including Piecewise Aggregate 
Approximation (PAA)\cite{Keogh2001Dimensionality,Guo2010An}, 
Symbolic Aggregate approximation (SAX)\cite{Lin2003A,Sun2014An}. PAA is 
competitive with or faster than other methods and it is easy to implement, 
which allows more flexible distance measure.
However, PAA algorithm is easy to lose an 
amount of information, especially the trend. For instance, if two series have 
same mean but opposite trend, PAA will judge these two sequences similar.  
Guo C \cite{Guo2010An} proposed an approach with PAA based on 
variance feature, which including forms of linear and square root,
to add some important information and solve the problem of same mean value.
While Sun Y \cite{Sun2014An} tried to add 
trend information by using starting and ending points of segments, 
whereas the starting and ending points do not reflect the trend in many case 
like the situation that both points have same value while the trend in segment 
are different.

In this paper, we propose two new approaches for time series similarity 
measurement that utilize trend feature. Our first method divides 
each segment into two parts based on mean value and use the 
numerical average  respectively to 
represent the trend. And we prove our method satisfies lower bound which 
guarantee no false dismissals. Our second method uses a binary string to record 
the trend 
change of a time series. The trend distance between two sequence 
is added to the PAA distance as the final distance to measure the similarity in 
both measures.

The remainder of the paper is organized as follows: Section 2 provides the 
background knowledge of original PAA and its limitations in detail. Section 3 
presents our proposed method and explain the trend representation. Section 4 
presents the experimental results of classification and anomaly detection on 
several data sets. Finally, section 5 concludes the paper.

\section{Background}

Piecewise Aggregate Approximation (PAA) is an approach of average dimensional 
reduction, which divides the time sequence equally and take the mean value of 
each segment as representation. Given time series, $Q=\{q_1,...,q_n\}$ , 
it will be reduced to a vector of length $w$ and presents as $\bar 
Q=\{\bar q_1,\bar q_2..., \bar q_w\}$ , where $w \le n$, the $i$th element of 
$\bar q$ is calculated by:
\begin{equation}
	\bar q_i=\frac{w}n \sum 
	\limits^{\frac{n}wi}_{j=\frac{n}w(i-1)+1}q_j\label{eq1}
\end{equation}
where the 
time series is divide into $w$ equi-size segments, $\bar q_i$ is the mean value 
of the $ith$ segment, $q_j$ is one of the 
time point in its segment(in which $ i \in w $ and $ j \in [q_{i},q_{i+1}] $ ). 

\begin{figure} \centering 
  	\subfigure[segment $w=2$.] { \label{fig1:a} 
  		\includegraphics[width=0.46\textwidth]{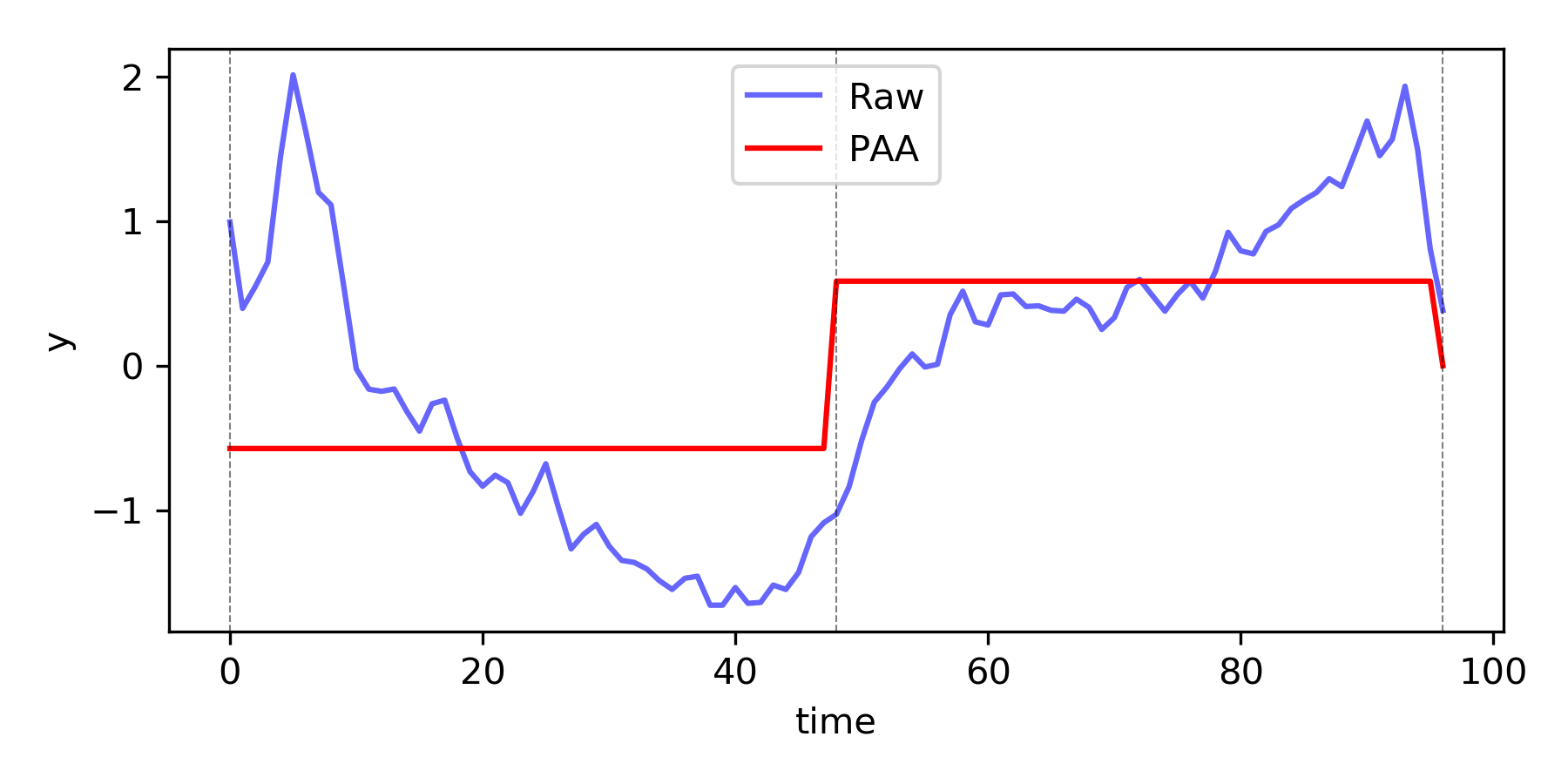} 
  	} 
  	\subfigure[segment $w=8$.] { \label{fig1:b} 
  		\includegraphics[width=0.46\textwidth]{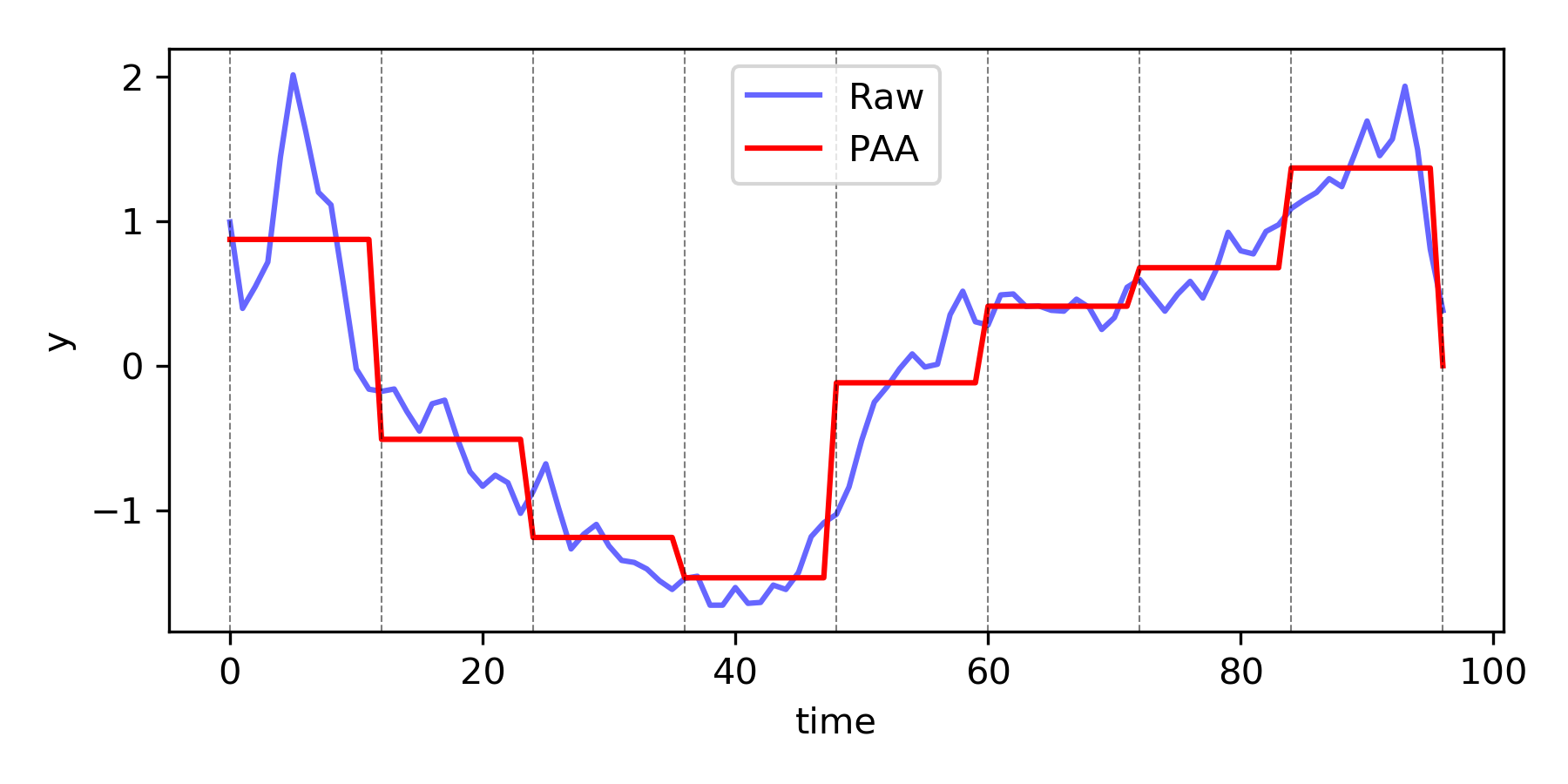} 
  	} 
  	\subfigure[segment $w=16$.] { \label{fig1:c} 
  		\includegraphics[width=0.46\textwidth]{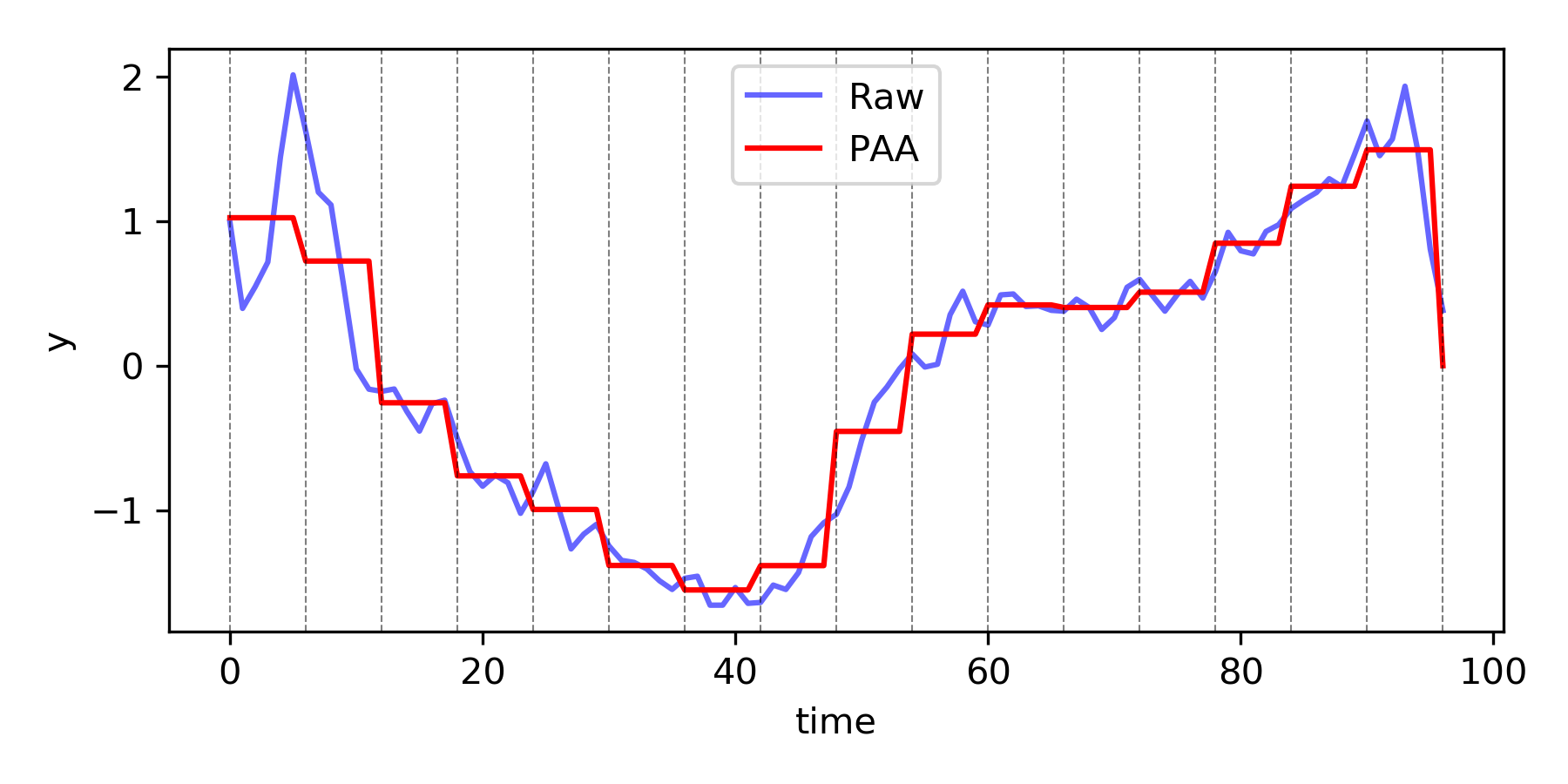} 
  	} 
  	\subfigure[segment $w=24$.] { \label{fig1:d} 
  		\includegraphics[width=0.46\textwidth]{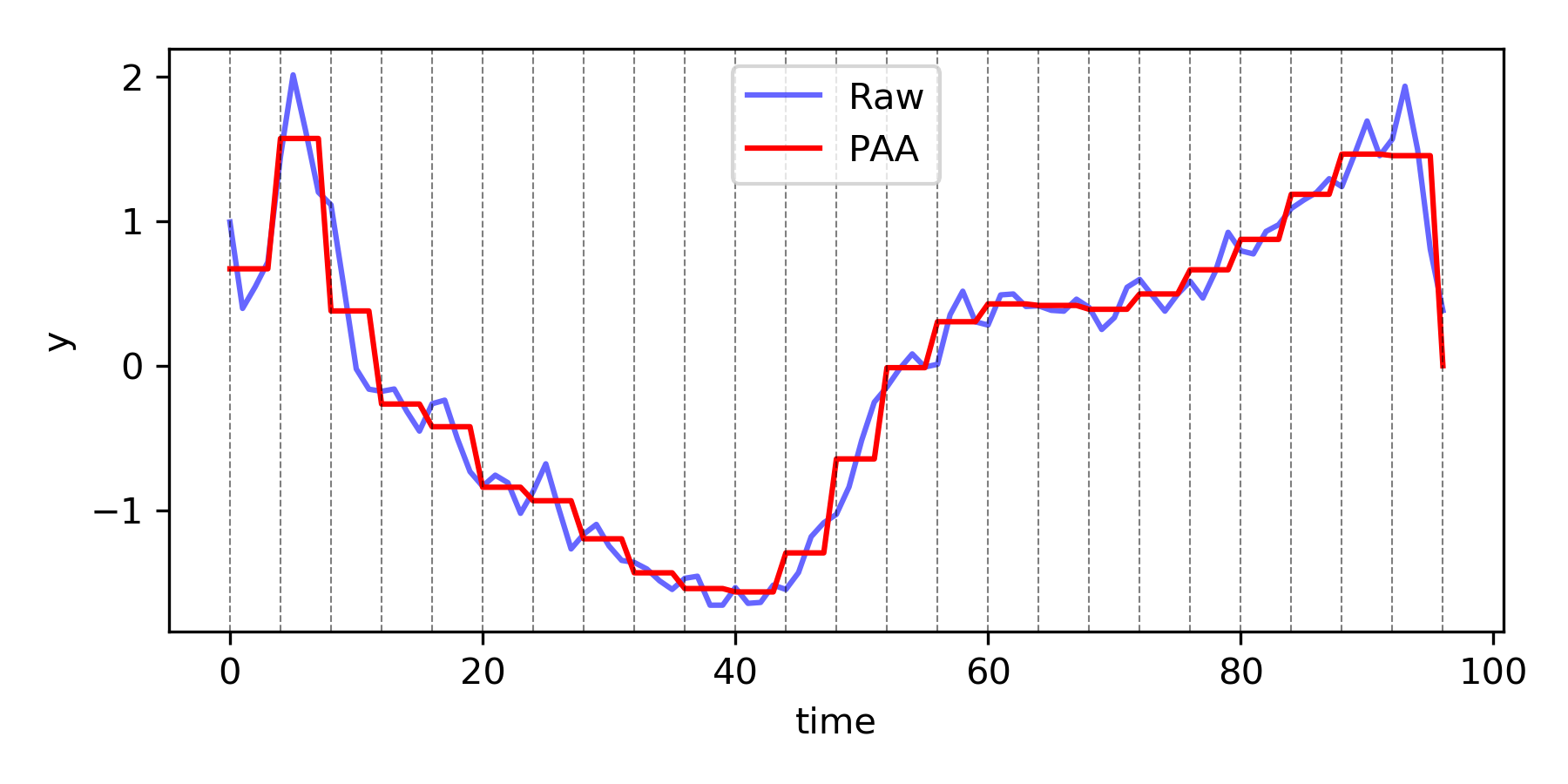} 
  	} 
  	\caption{ PAA representation for one of the time series in ecg200. In this 
  	case, \textit{Raw} 
  		represents for the 
  		original time series and \textit{PAA} represents for the transformed 
  		time 
  		series, 
  		and 
  		the length of the time series is 96.} 
  	\label{fig1} 
  \end{figure}

The mean value of data falling within the segment will be calculated and the 
mean value will replace whole 
segment as the new representation. Once the length of segment becomes larger, 
it will lead to the loss of trend information as shown in Fig.\ref{fig1}. To 
illustrate this point, Fig.\ref{fig1:a} divide the time series into two 
subsequences, the length is $96/2=48$, which lose the trend information a lot 
and is obviously quite different from the original one. With the increase of 
the number of segments in Fig.\ref{fig1:b} to Fig.\ref{fig1:d}, the reduced 
dimension sequence is closer to the original one.
Besides, the result of reduction will be inaccurate when two sequence have same 
mean value while the trends are different. For PAA method, the distance measure 
was proposed as Equation(\ref{eq2}).

\begin{equation}
	Dist(\bar Q,\bar P)
	=\sqrt{\frac{n}{w}}\sqrt{\sum_{i=1}^{w}(\bar{p_i}-\bar{q_i})^2}\label{eq2}
\end{equation}
\begin{equation}
ED(Q,P)
=\sqrt{\sum_{i=1}^{n}(p_i-q_i)^2}\label{eq3}
\end{equation}

\begin{figure}
	\begin{center}
	\includegraphics[width=0.7\textwidth]{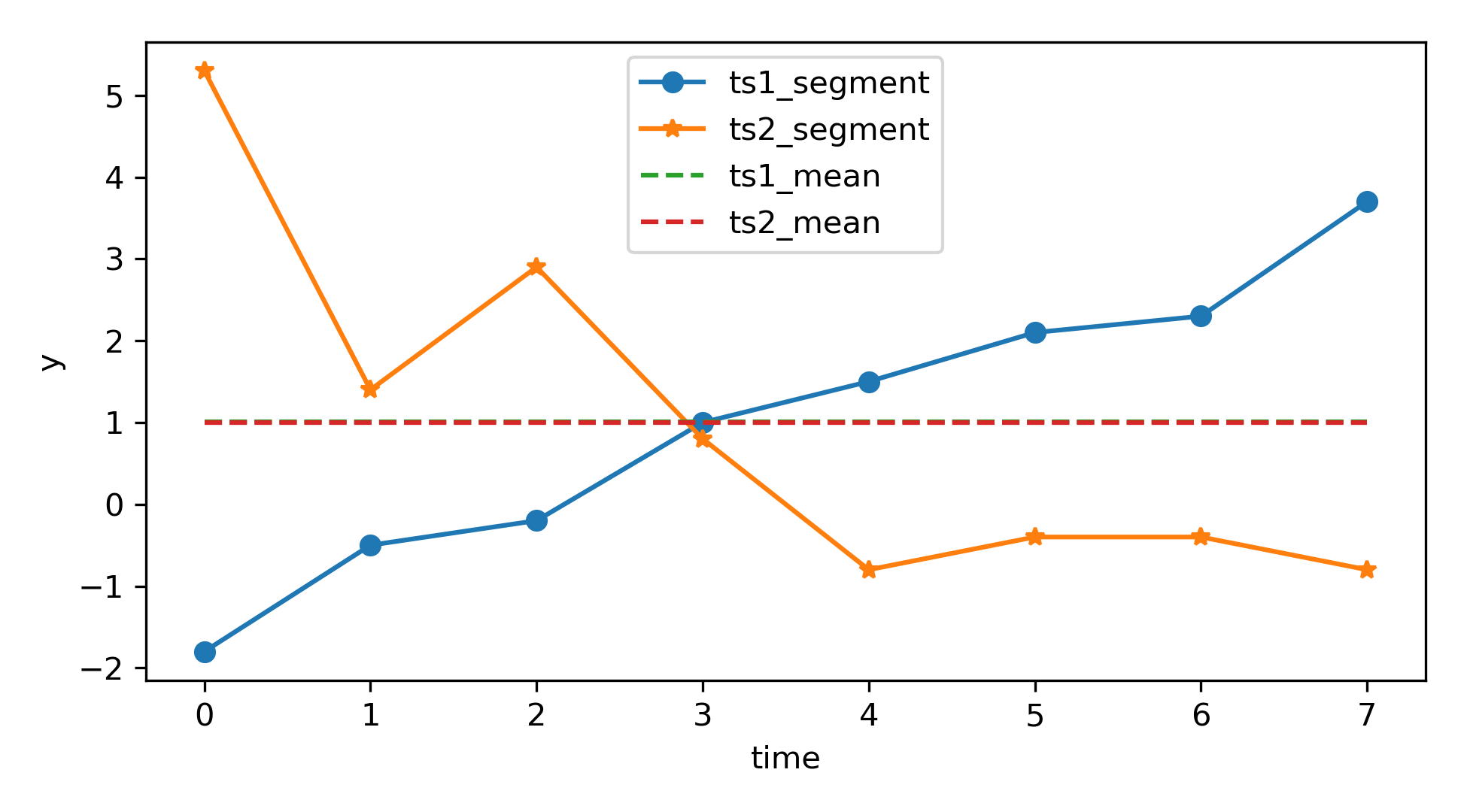}
	\caption{The comparison of one segment in two time series which have very 
	close mean value, the mean value of ts1 is 1.012 and ts2 is 1.01. To 
	illustrate that t1 and t2 are different
	while through the PAA distance calculation, they are similar.} \label{fig2}
\end{center}
\end{figure}

Compared with the original Euclidean distance in Equation(\ref{eq3}) which is 
one of the true distance measures\cite{Rabiner1993Fundamentals}. It can be 
seen from the above two formulas that Euclidean distance calculates every time 
points' distance while the PAA reduces the n dimension to w dimension and 
simply 
multiple $n/w$ to enlarge time series, which roughly cover up information in 
detail. To illustrate this, Fig.\ref{fig2} shows that ts1 and ts2 are the 
relative segment in two time series. Even if it can be seen intuitively from 
the 
figure that they are two different time series, but their mean values are 
very close and the distance measure calculated by PAA will obtain that the two 
time series are similar.

Furthermore, PAA is an approximate method to fit the original sequence, so the 
maximum and minimum value will be missed. 
To address above problems, we propose two methods to record the trend 
information, and the detail methods 
will be describe in section 3.

\section{Our Proposed Methods }

As we review above, we know that the original PAA method simply flattens the
curve by segments, which will lose a lot of information, especially the trend
change information. For the propose of solving this problem, we propose two 
methods.
The first method is based on relative mean value of each segment to 
record the trend, and we call it Numerical Trend Based On PAA(NT\_PAA). It 
divides each segment into two parts by mean value and calculate the numerical 
average 
of them separately, with the trend distance calculated by difference. The 
other is a method of recording the relative trend change for a sequence by 
using a binary string, we name it Binary Trend Based On 
PAA(BT\_PAA). And the trend distance is the number of different binary strings 
in 
two time series, which is weighted by the number of segments. Both the 
final distance combine the trend approximate distance with the PAA distance.

\subsection{Numerical Trend Based On PAA}
\subsubsection{The trend representation}
For the propose of solving this problem, we add the incremental representation 
by using the numerical mean value on behalf of trend to improve it.
A time series of length $n$ represents as $Q=\{q_1,...,q_n\}$ which is 
divided into $w$ segments, $Q=\{\bar{q_1},...,\bar{q_w}\}$, the 
formula is shown 
in Equation.\ref{eq1} 

\begin{figure}
	\begin{center}
	
	\includegraphics[width=0.7\textwidth]{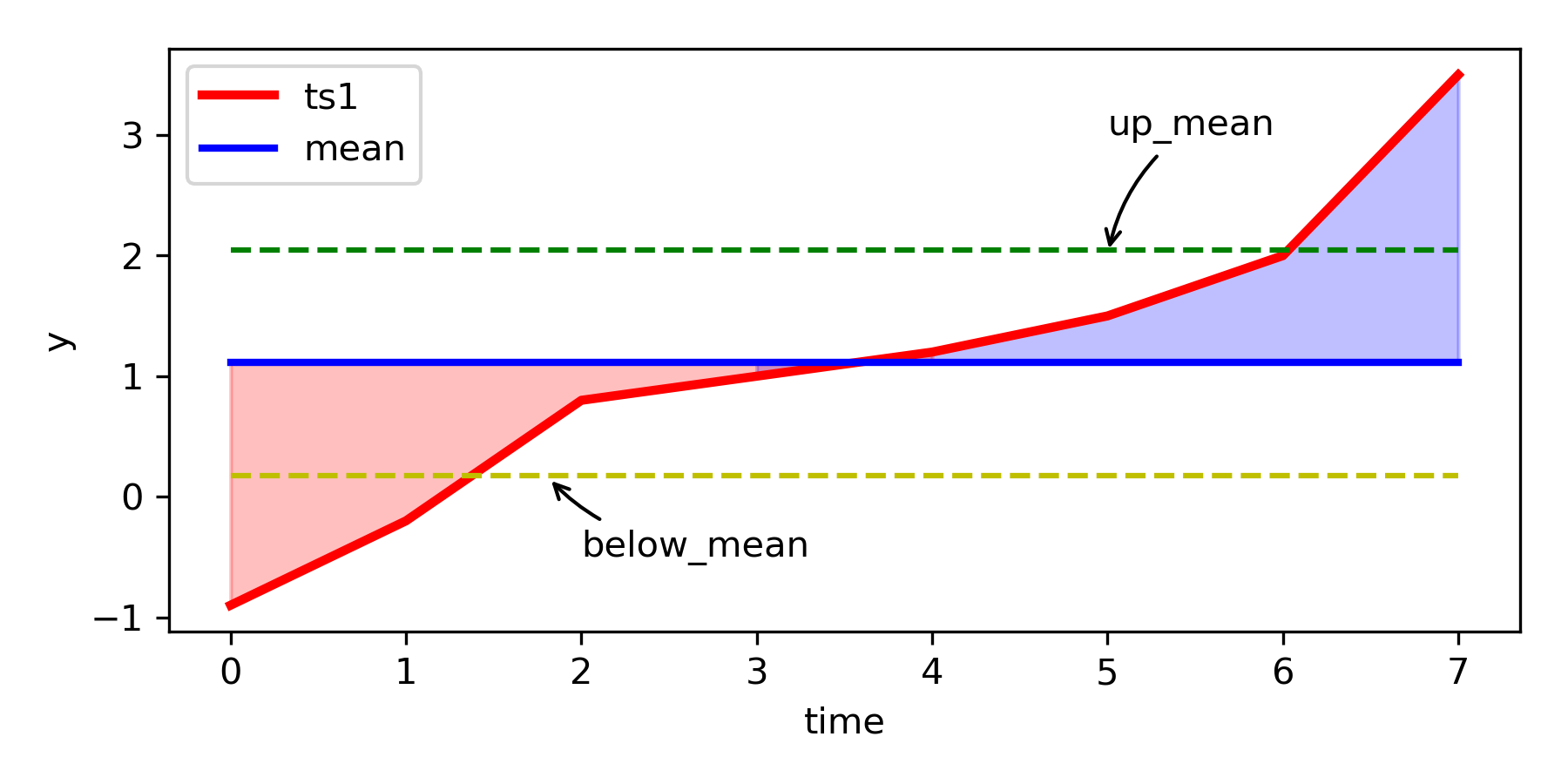}
	\caption{\textit{up\_mean} and \textit{below\_mean} in one time series 
	segment.} \label{fig4}
\end{center}
\end{figure}

We define \emph{up difference} as the difference of all time points in one 
segment 
above the mean value, 
while \emph{below difference} as the difference of all the 
time 
points in one segment below the mean value.
Therefore the up-mean value $\Delta q_u$ and below-mean value $\Delta q_b$ 
which are relative 
to the mean value in each segment can be defined as:
\begin{equation}
\Delta q_{ui}=\frac{1}{u_i}\sum \limits^{\frac{n}wi}_{k=\frac{n}w(i-1)+1}(q_k- 
\bar 
q_i), (q_k 
\ge \bar q_i)
\label{eq_u}
\end{equation}

\begin{equation}
\Delta q_{bi}=\frac{1}{b_i}\sum \limits^{\frac{n}wi}_{k=\frac{n}w(i-1)+1}(\bar 
q_i-q_k), (q_k 
< \bar q_i)
\label{eq_b}
\end{equation}
where $u_i$ is the number of up value in $ith$ segment, and $b_i$ is the number 
of below value , 
$\frac{w}{n}=u_i+b_i$, and we can see from the Fig.\ref{fig4} clearly that the 
time points in red area is the below difference value and time points in blue 
area is the up difference value. 

\subsubsection{Distance Measure}
In order to guarantee no false dismissals, we must produce a distance measure 
defined in index space. We can define the trend distance based on numerical 
mean value in one segment 
as follows.

	\begin{equation}
	nt(q,c)=\sqrt{{u(\Delta q_{u} - \Delta c_{u})^2}+b(\Delta q_{b}-\Delta  
	c_{b})^2 }
\label{eq_nt}
	\end{equation}

And the final distance between two time series based on trend approximation can 
be 
defined as:

\begin{equation}
NT\_Dist(Q,C)
=\sqrt{\frac{n}w\sum\limits_{i = 1}^w {(\bar q_i- \bar c_i)^2}+\sum\limits_{i 
= 1}^wnt(q_i,c_i)^2} \\
\label{eq_ntd}
\end{equation}

Our proposed method is a lower bounding measure to Euclidean Distance(ED) which 
can be proved as follow.
\begin{proof}
	According to the\cite{Keogh2001Dimensionality}, the authors have already 
	proved that the PAA distance is lower bound the Euclidean distance:
	\begin{equation}
	ED \ge \sqrt{\frac{n}w\sum\limits_{i = 1}^w {{{({\bar q_i} - {\bar 
	c_i})}^2}}} 
\label{eq_eu}
	\end{equation}
	
	In order to prove the $NT\_Dist(Q,C)$ lower bounds Euclidean Distance, we should expand the Euclidean first, where $q_i$ can be represented as $q_i=\bar q_i - \Delta q_i$ , so as $c_i=\bar c_i - \Delta c_i$, and simply make the $w=1$.
	
	\begin{equation}
	\begin{split}
		ED^2
		&=\sum\limits_{i = 1}^n {{{({q_i} - {c_i})}^2}} \\
		&=\sum\limits_{i = 1}^n((\bar q_i - \Delta q_i)-(\bar c_i - \Delta 
		c_i))^2 \\
		&=n(\bar q_i -\bar c_i )^2+2(\bar q_i -\bar c_i )\sum\limits_{i = 1}^n( \Delta q_i - \Delta c_i)+\sum\limits_{i = 1}^n( \Delta q_i - \Delta c_i)^2
	\end{split}
	\label{eq9}
	\end{equation}
	
	We already know that $(\bar q_i -\bar c_i )\sum\limits_{i = 1}^n( \Delta 
	q_i - \Delta c_i)=0$, therefore, ED can be transformed as follows:
	\begin{equation}
	ED^2=n(\bar q_i -\bar c_i )^2+\sum\limits_{i = 1}^n( \Delta q_i - \Delta 
	c_i)^2
	\label{eq10}
	\end{equation}
	And for our method, we can expand our method from Equation(9) that
	\begin{equation}
		Dist^2=n({\bar q_i} - {\bar c_i})^2+NT(q_i,c_i)^2
		\label{eq11}
	\end{equation}
	Combine Equation(12) and (13), we only have to prove that 
	\begin{equation}
	\sum\limits_{i = 1}^n( \Delta q_i - \Delta c_i)^2 \ge {u(\Delta q_{u} - 
	\Delta c_{u})^2}+b(\Delta q_{b}-\Delta  
	c_{b})^2
	\label{eq12}
	\end{equation}
	
	Equation(12) can be divided into two parts including \textit{up} and 
	\textit{below} 
	area as we mention above due to $\frac{n}{w}=u+b$, 
	\begin{equation}
		\sum\limits_{i = 1}^u( \Delta q_i - \Delta c_i)^2 \ge u(\Delta q_{u} - 
		\Delta c_{u})^2
		\label{eq13}
	\end{equation}
	\begin{equation}
	\sum\limits_{i = 1}^b(\Delta q_i - \Delta c_i)^2 \ge b(\Delta q_{b}-\Delta  
	c_{b})^2
	\label{eq14}
	\end{equation}
	where $\Delta q_u$ and $\Delta q_b$ are the mean value in different two 
	parts, which can be defined as Equation(4) and (5). 
	In other words, it can be represented as $\Delta q_i=\Delta 
	q_{ui}-\Delta(\Delta q_i)$ and $\Delta c_i=\Delta c_{ui}-\Delta(\Delta 
	c_i)$.
	To prove Equation(13) and (14), the process are the same as Equation(8). 
	The 
	prove is done.

\end{proof}

\subsection{Binary Trend Based On PAA }

\subsubsection{The trend representation}

Another method to represent the trend is based on binary string, which can roughly but efficiently reflect the relative trend change to mean value in each segment. 
We can use binary string $B=\{0,1\}^n$ to represent the trend relative to the 
mean and the bits are defined as follow:

\begin{equation}
b_j=\left\{
\begin{array}{lr}
1, & p_j \ge \bar  p_i\\
0, & p_j < \bar  p_i
\end{array}
\right. 
\label{eq15}
\end{equation}
in Equation(\ref{eq15}), each raw data point segment is represented as 
\textbf{\textit{1}} when 
the raw 
data is greater than the mean value of $ith$ segment, otherwise, if the raw 
data is less than the mean, it is represented as \textbf{\textit{0}}. 

\begin{figure} \centering 
	\subfigure[time series $ P_i $.] { \label{fig:a} 
		\includegraphics[width=0.46\textwidth]{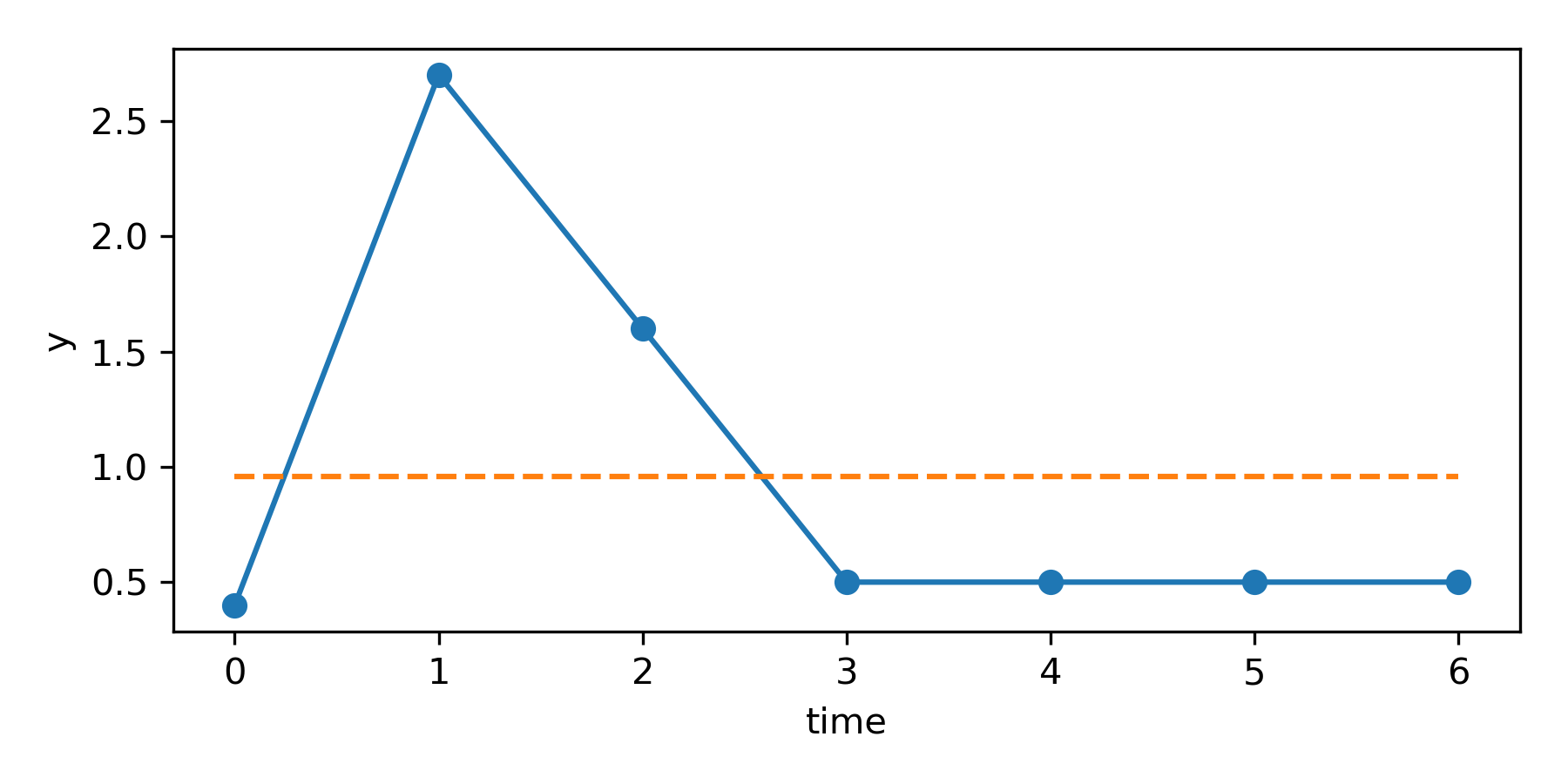} 
	} 
	\subfigure[time series $ Q_i $.] { \label{fig:b} 
		\includegraphics[width=0.46\textwidth]{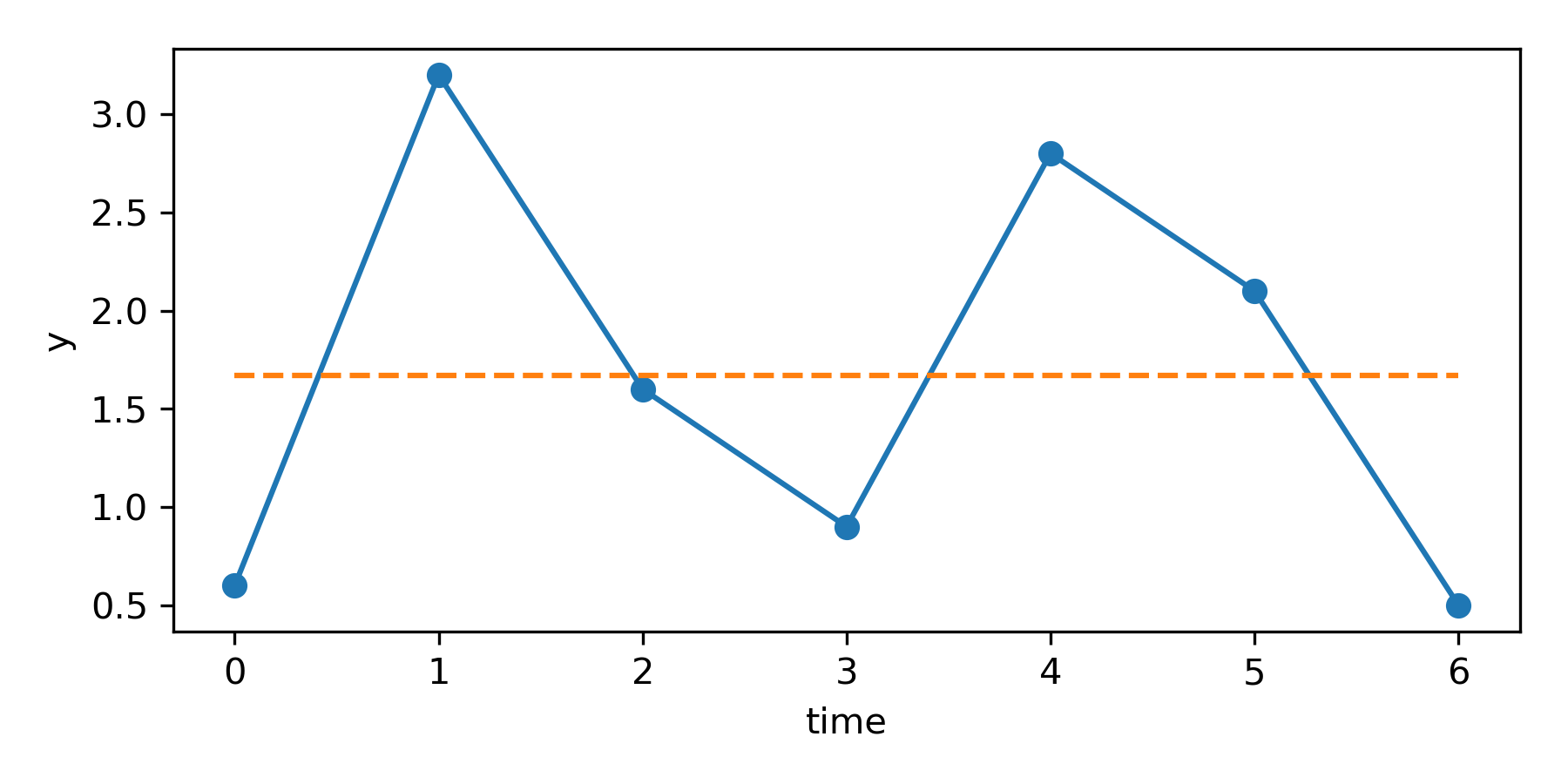} 
	} 
	\caption{ The trend representation. $P_i$ and $Q_i$ represent for one of 
	the corresponding segment in time series $P$ and $Q$, and the dotted line 
	is the mean value. In \ref{fig:a}, the mean value is 0.8375 and binary 
	string is $B_{Pi}=0110000$, in \ref{fig:b}, the mean value is 1.6714 and 
	binary string is $B_{Qi}=0100110$.} 
	\label{fig3} 
\end{figure}

For example, suppose we have one of the corresponding segment in two time 
series $P_i$ and $Q_i$, as we can see in Fig.\ref{fig3} 
$$P_i=\{0.4,2.7,1.6,0.5,0.5,0.5,0.5\}$$  
$$Q_i=\{0.6,3.2,1.6,0.9,2.8,2.1,0.5\}$$ so we can calculate the mean value as 
$mean(P_i)=0.8375$ and $mean(Q_i)=1.6714$, then compare each raw time point 
with mean value, we can get the binary string as $B_{Pi}=0110000$,  and 
$B_{Qi}=0100110$. 

\subsubsection{Distance Measure}

The trend distance of the binary string between two series is as follows, where 
the length of time series is $ n$ and it is divided into $w$ segment.
\begin{equation}
	\begin{split}
	bt(\bar Q,\bar P)
	&=\sqrt{\sum_{i=1}^w\frac{w}{n}count(b_{pi} \oplus b_{qi})} \\
	&=\sqrt{\frac{w}{n}count(B_P \oplus B_Q)}
	\end{split}
	\label{eq_bt}
\end{equation}
 $b_{pi} , b_{qi}$ are the binary string of corresponding segment of two 
 series, and the function $count$ is used to sum up the number of 1 in the 
 binary string. The formula can be transformed where $B_P$ and $B_Q$ are the 
 whole binary strings of two time series. 

Finally, we can define the BIT\_Dist measure function based on trend distance 
and PAA as follows, 
\begin{equation}
BIT\_Dist(\bar Q,\bar 
P)=\sqrt{\frac{n}{w}}\sqrt{\sum_{i=1}^{w}(\bar{p_i}-\bar{q_i})^2}+\sqrt{\frac{w}{n}count(B_P
 \oplus B_Q)} 
\label{eq_btd}
\end{equation}

From Equation(\ref{eq_btd}), it can be seen that the effect of trend distance 
on 
the 
overall distance is weighted by $w/n$, which $n$ is fixed. The larger of $w$, 
the greater the proportion of trend distance and the longer length of 
one segment. Once the subsequence is very long, the trend among this segment 
will change into a parallel line with no trend change, therefore, the increase 
of trend distance helps distinguish between the similarity of two subsequence. 
On the contrary, the smaller of $w$, the smaller proportion of trend distance. 
Because if the length of subsequence is small, even contains only two time 
points, their trend is similar to linear, which will not lose trend information 
a lot.

\section{Experiments}
In this section, we evaluate our proposed methods and present the results of 
experiments. First, we introduce the data sets we used in experiments. Then we 
compare the 
performance of proposed methods in aspect of classification and anomaly 
detection. The experiments are performed on 2.5GHz processor with 16GB physical 
memory. We use cross-validation to find the optimal reduction ratio $ s=n/w $ 
on the training data sets and verify them on the verification data sets.

\subsection{Dataset}

We perform all the experiments over the UCR Time Series Classification Archive 
repository\cite{UCRArchive}, which is a large and mature open data sets, and 
each of the datasets is divided into a training data set and a test data set. 
We 
choose 24 data sets in UCR and the classes of time series are between 2 and 39, 
the length of time series are between 84 and 1024 with the total size of the 
data sets are between 60 and 2000. The detail of data sets is shown in 
Table \ref{tab1}

\begin{table}[t]
	\newcommand{\tabincell}[2]{\begin{tabular}{@{}#1@{}}#2\end{tabular}}
	\centering
	\caption{The description of time series data sets we used(from UCR Time 
	Series Classification Archive repository).}
	\label{table:data_set}
	\begin{tabular}{lcrrrr}
		\toprule
		No. & data sets &  classes & \tabincell{c}{Size of\\ training set} & \tabincell{c}{Size of\\ testing set} & \tabincell{c}{Length of\\ times series}\\
		\midrule
		1&Adiac & 37    & 390   & 391   & 176 \\
		2&Beef  & 5     & 30    & 30    & 470 \\
		3&Car   & 4     & 60    & 60    & 577 \\
		4&Coffee & 2     & 28    & 28    & 286 \\
		5&Computers & 2     & 250   & 250   & 720 \\
		6&Earthquakes & 2     & 139   & 322   & 512 \\
		7&ECG200 & 2     & 100   & 100   & 96 \\
		8&ECGFiveDays & 2     & 23    & 861   & 136 \\
		9&FaceFour & 4     & 24    & 88    & 350 \\
		10&FISH  & 7     & 175   & 175   & 463 \\
		11&Gun\_Point & 2     & 50    & 150   & 150 \\
		12&Ham   & 2     & 109   & 105   & 431 \\
		13&Herring & 2     & 64    & 64    & 512 \\
		14&Lighting2 & 2     & 60    & 61    & 637 \\
		15&MoteStrain & 2     & 20    & 1252  & 84 \\
		16&OSULeaf & 6     & 200   & 242   & 427 \\
		17&Phoneme & 39    & 214   & 1896  & 1024 \\
		18&Plane & 7     & 105   & 105   & 144 \\
		19&ShapeletSim & 2     & 20    & 180   & 500 \\
		20&Strawberry & 2     & 370   & 613   & 235 \\
		21&SwedishLeaf & 15    & 500   & 625   & 128 \\
		22&ToeSegmentation2 & 2 & 36 & 130 & 343 \\
		23&Trace & 4     & 100   & 100   & 275 \\
		24&Wine             & 2 & 57 & 54  & 234 \\
		\bottomrule
	\end{tabular}%
	\label{tab1}%
\end{table}%

\subsection{Experimental setup}
\subsubsection{Method:}
Since our method is to improve the PAA based on trend feature, we compare the 
accuracy and effectiveness on classification and anomaly detection with  
Piecewise Aggregate Approximation(PAA), Euclidean Distance(ED) and Cosin 
Similarity(CO) distance measures. Cosin similarity\cite{Chomboon2015An} uses 
the 
cosin of the angle between two vectors in vector space as the measure of the 
difference within two individuals. As for distance representation, it shall be 
$1$ minus the cosin similarity distance in our experiment.
For the classification process,  we conduct the experiments using the k-Nearest 
Neighbor (K-NN) classifier and set the k=3, of which the accuracy is determined 
by the similarity distance between test sample and each of training data. And 
for the anomaly detection process, we use Local Outlier Factor 
(LOF)\cite{Breunig2000LOF} to look for the optimal parameters of nearest 
neighbor 
in test sets by cross-validation. 

\subsubsection{Evaluation metrics:}

In these experiments, error rate, precision, recall and F1-score are used as 
evaluation metrics to evaluate the performance of classification. Precision is 
how much of the retrieved entries is accurate, while recall is how many 
accurate entries have been retrieved. As for F1-score, it is the harmonic 
average of precision and 
Recall. When F1 is higher, 
the comparison shows that the experimental method is ideal.
\begin{equation}
	Error\ rate=\frac{Number\ of\ incorrect\ classification}{Total\ number\ of\ 
	test\ samples}
\end{equation}
\begin{equation}
	F1=2*\frac{precision8recall}{presion+recall}
\end{equation}

We use Area Under Curve (AUC) evaluation metric to measure the performance of 
anomaly detection. AUC is defined as the area under the ROC curve and the range 
of values is between 0.5 and 1. The greater the AUC value, the better the 
detection algorithm effect.

\begin{table}[t]
	\centering
	\caption{The result of 3-NN classification for NT\_PAA and BT\_PAA. s 
	represents the best number of points in a time series segment. The highest 
	values are highlighted in bold.}
	\begin{tabular}{c|cccccccccccccc}
		\toprule
		\multicolumn{1}{c|}{\multirow{2}[4]{*}{data set}} & 
		\multicolumn{3}{c}{BT\_PAA} & \multicolumn{3}{c}{NT\_PAA} & 
		\multicolumn{3}{c}{PAA} & \multicolumn{3}{c}{Cosin} & 
		\multicolumn{2}{c}{ED} \\
		\cmidrule{2-15}    \multicolumn{1}{c|}{} & s     & error & F1    & 
		s     & error & F1    & s     & error & F1    & s     & error & F1    & 
		error & F1 \\
		\midrule
		1     & 15    & \textbf{0.014} & 0.821 & 6     & 0.015 & \textbf{0.886} 
		& 2     & 0.018 & 0.814 & 3     & 0.018 & 0.814 & 0.023 & 0.814 \\
		2     & 2     & \textbf{0.017} & \textbf{0.973} & 11    & 0.033 & 0.944 
		& 2     & 0.067 & 0.822 & 3     & 0.067 & 0.880 & 0.133 & 0.822 \\
		3     & 11    & \textbf{0.067} & \textbf{0.897} & 8     & 0.133 & 0.780 
		& 18    & 0.083 & 0.893 & 18    & 0.083 & 0.893 & 0.158 & 0.813 \\
		4     & 3     & \textbf{0.000} & \textbf{1.000} & 18    & 0.000 & 0.982 
		& 2     & 0.018 & 0.982 & 2     & 0.018 & 0.982 & 0.018 & 0.982 \\
		5     & 2     & 0.396 & 0.601 & 9     & \textbf{0.328} & \textbf{0.693} 
		& 2     & 0.406 & 0.594 & 4     & 0.404 & 0.583 & 0.410 & 0.589 \\
		6     & 17    & 0.215 & 0.555 & 18    & \textbf{0.200} & \textbf{0.576} 
		& 15    & 0.226 & 0.547 & 15    & 0.210 & 0.556 & 0.265 & 0.474 \\
		7     & 7     & \textbf{0.085} & \textbf{0.901} & 2     & 0.100 & 0.880 
		& 3     & 0.090 & 0.896 & 3     & 0.090 & 0.556 & 0.090 & 0.896 \\
		8     & 5     & \textbf{0.001} & \textbf{0.999} & 8     & 0.019 & 0.973 
		& 3     & 0.002 & 0.998 & 3     & 0.005 & 0.997 & 0.009 & 0.991 \\
		9     & 19    & \textbf{0.009} & 0.947 & 15    & 0.027 & 0.958 & 19    
		& 0.018 & 0.947 & 14    & 0.018 & \textbf{0.972} & 0.027 & 0.958 \\
		10    & 15    & 0.080 & 0.811 & 2     & 0.080 & \textbf{0.825} & 2     
		& 0.080 & 0.824 & 16    & 0.080 & 0.825 & 0.089 & 0.824 \\
		11    & 4     & \textbf{0.040} & 0.960 & 4     & 0.050 & \textbf{0.965} 
		& 7     & 0.045 & 0.955 & 7     & 0.045 & 0.955 & 0.050 & 0.950 \\
		12    & 13    & \textbf{0.173} & \textbf{0.826} & 19    & 0.248 & 0.743 
		& 9     & 0.178 & 0.820 & 13    & 0.159 & 0.816 & 0.206 & 0.791 \\
		13    & 6     & 0.453 & 0.527 & 10    & \textbf{0.336} & \textbf{0.611} 
		& 15    & 0.477 & 0.505 & 17    & 0.484 & 0.505 & 0.500 & 0.482 \\
		14    & 19    & 0.190 & \textbf{0.789} & 13    & \textbf{0.174} & 0.799 
		& 17    & 0.198 & 0.784 & 17    & 0.207 & 0.784 & 0.248 & 0.729 \\
		15    & 8     & \textbf{0.038} & \textbf{0.962} & 6     & 0.046 & 0.949 
		& 8     & 0.045 & 0.955 & 5     & 0.207 & 0.951 & 0.081 & 0.918 \\
		16    & 2     & 0.045 & 0.909 & 17    & \textbf{0.041} & \textbf{0.901} 
		& 2     & 0.045 & 0.903 & 9     & 0.045 & 0.903 & 0.048 & 0.903 \\
		17    & 17    & 0.020 & 0.528 & 4     & \textbf{0.019} & \textbf{0.541} 
		& 17    & 0.020 & 0.527 & 3     & 0.020 & 0.534 & 0.033 & 0.506 \\
		18    & 19    & 0.019 & 0.961 & 19    & \textbf{0.014} & \textbf{0.970} 
		& 14    & 0.019 & 0.952 & 16    & 0.019 & 0.960 & 0.033 & 0.933 \\
		19    & 2     & 0.400 & 0.548 & 19    & 0.455 & 0.560 & 19    & 
		\textbf{0.395} & \textbf{0.558} & 4     & 0.295 & 0.547 & 0.460 & 0.536 
		\\
		20    & 19    & \textbf{0.035} & \textbf{0.958} & 12    & 0.055 & 0.936 
		& 10    & 0.044 & 0.953 & 10    & 0.044 & 0.953 & 0.051 & 0.945 \\
		21    & 14    & 0.031 & \textbf{0.856} & 18    & 0.036 & 0.844 & 18    
		& \textbf{0.030} & 0.863 & 18    & 0.033 & 0.863 & 0.047 & 0.791 \\
		22    & 8     & 0.120 & 0.823 & 15    & \textbf{0.108} & \textbf{0.836} 
		& 9     
		& 0.120 & 0.823 & 12    & 0.114 & 0.823 & 0.133 & 0.805 \\
		23    & 15    & 0.020 & 0.937 & 15    & 0.045 & 0.991 & 2     & 0.020 & 
		\textbf{1.000} & 2     & \textbf{0.000} & 1.000 & 0.055 & 1.000 \\
		24    & 16    & 0.063 & 0.974 & 14    & 0.000 & 0.924 & 16    & 0.000 & 
		0.973 & 4     & 0.035 & \textbf{0.980} & 0.000 & 0.921 \\
		\cline{1-15}
		Average &       & \textbf{0.105} & \textbf{0.836} &       & 0.107 & 
		\textbf{0.836} &       & 0.110 & 0.829 &       & 0.112 & 0.818 & 0.132 
		& 0.807 \\
		\bottomrule
	\end{tabular}%
	\label{tab2}%
\end{table}%

\subsection{Comparison in lower bound}
\begin{figure}
	\begin{center}
	\includegraphics[width=0.8\textwidth]{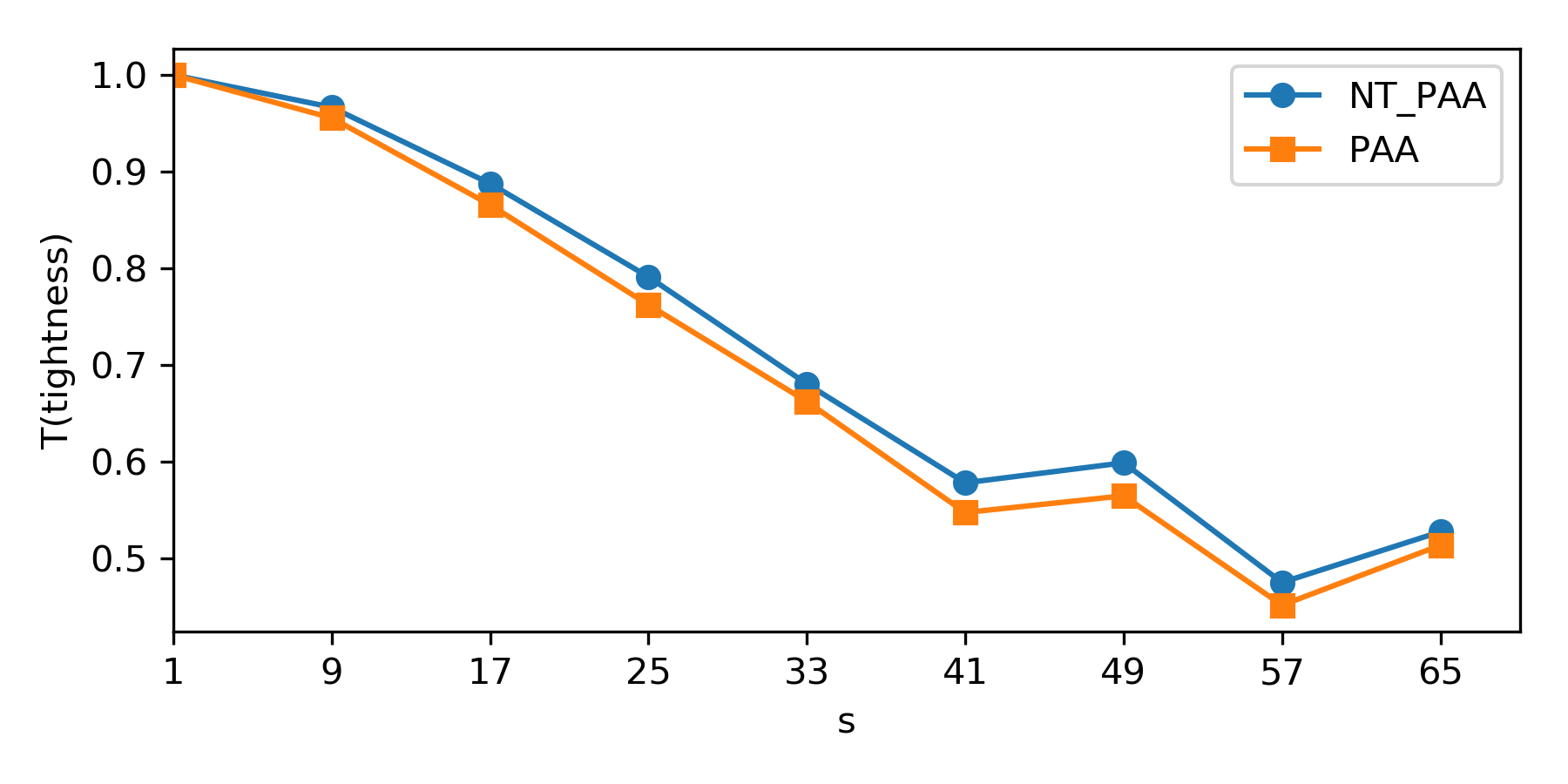}
	\caption{The comparison of tightness between NT\_PAA and PAA} \label{fig6}
\end{center}
\end{figure}

As for the NT\_PAA method, we already prove that our method has tighter 
lower bound than original PAA, which can be further proved by experiment as 
shown in Fig.\ref{fig6}. We choose the Euclidean distance as the true 
distance and make the tightness represent as Equation(20), where $Dist(P,Q)$ is 
the approximate distance measure, $s$ represent the reduction ratio, and T is 
range in $[0,1]$, the closer to 1 the better, 200 time series of length 150 are 
tested.
\begin{equation}
	T(tightness)=\frac{Dist(P,Q)}{ED(P,Q)}
\end{equation}

From this Fig.\ref{fig6} we can find that when the reduction ration is 1, 
the tightness is equal, and as the reduction ratio becomes bigger, the 
tightness 
becomes smaller.

\subsection{Comparison on Classification}

In this experiment, our proposed methods BT\_PAA and NT\_PAA are compared with 
three other distance measurements, Cosin\cite{Chomboon2015An}, 
Euclidean\cite{Faloutsos1994Fast} and PAA, with 24 data sets in UCR are used. 
The results of classification are shown in Table \ref{tab2}, and the best 
results are highlighted in bold font. 

To measure the improvement that the BT\_PAA and NT\_PAA classifier provide, the 
data sets are trained and tested with a varying window size $s$ comparing with 
original PAA, Cosin and ED. 
The results in Table \ref{tab2} show that all methods have different best 
number of $s$ ratio.
Furthermore, our proposed methods performs better than other 
three methods in most of the data sets in Table \ref{tab2}, BT\_PAA has most of 
lowest error rate 
in data sets(11/24) while NT\_PAA is less(8/24).  On the other hand, our 
methods 
perform almost the same in F1 metric. On average(10/24), our proposed methods
outperforms than PAA, Cosin and Euclidean under these evaluation metrics.

\begin{figure} \centering 
	\subfigure[] { \label{fig7:a} 
		\includegraphics[width=0.46\textwidth]{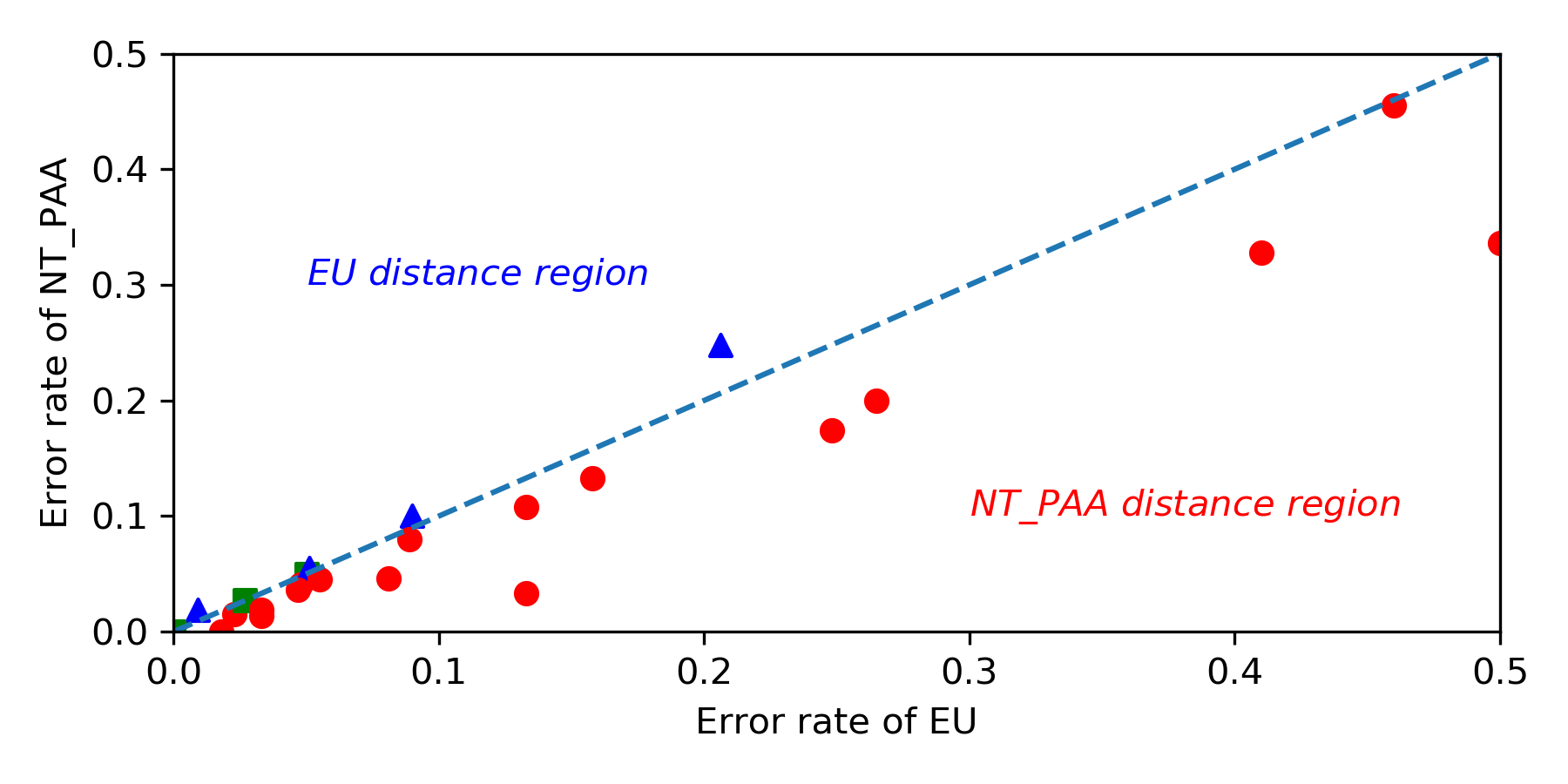} 
	} 
	\subfigure[] { \label{fig7:b} 
		\includegraphics[width=0.46\textwidth]{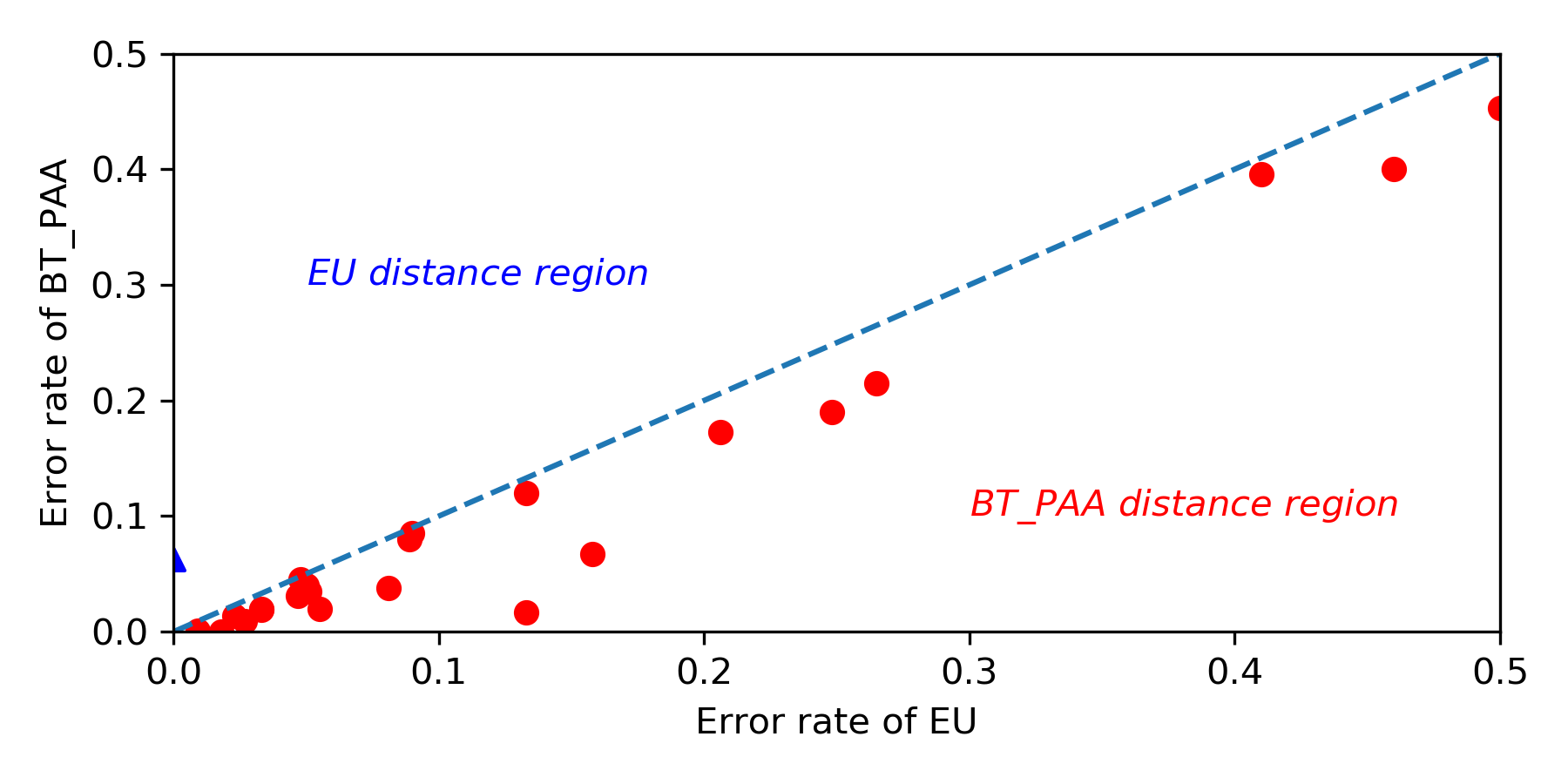} 
	} 
	\subfigure[] { \label{fig7:c} 
		\includegraphics[width=0.46\textwidth]{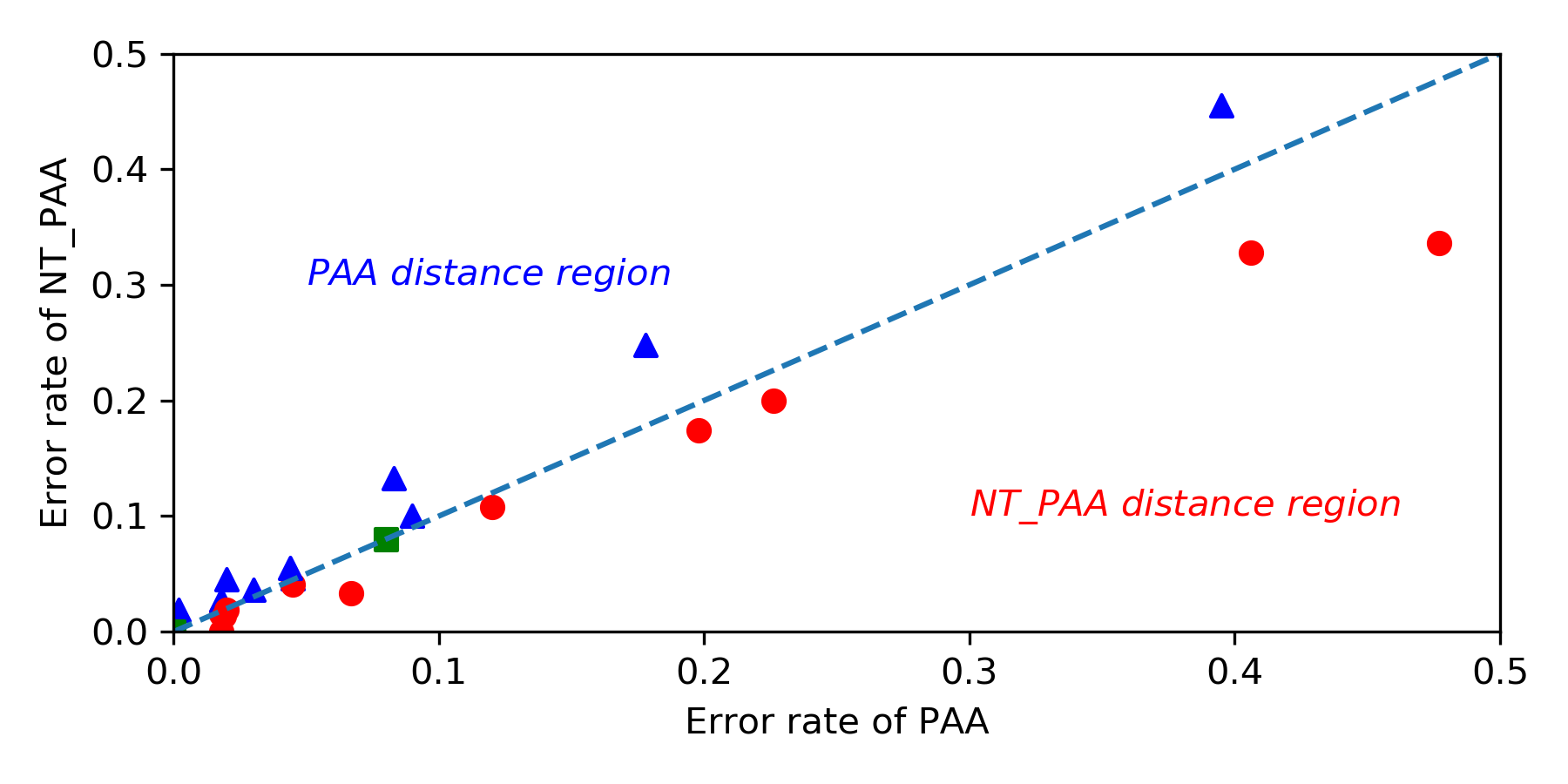} 
	} 
	\subfigure[] { \label{fig7:d} 
		\includegraphics[width=0.46\textwidth]{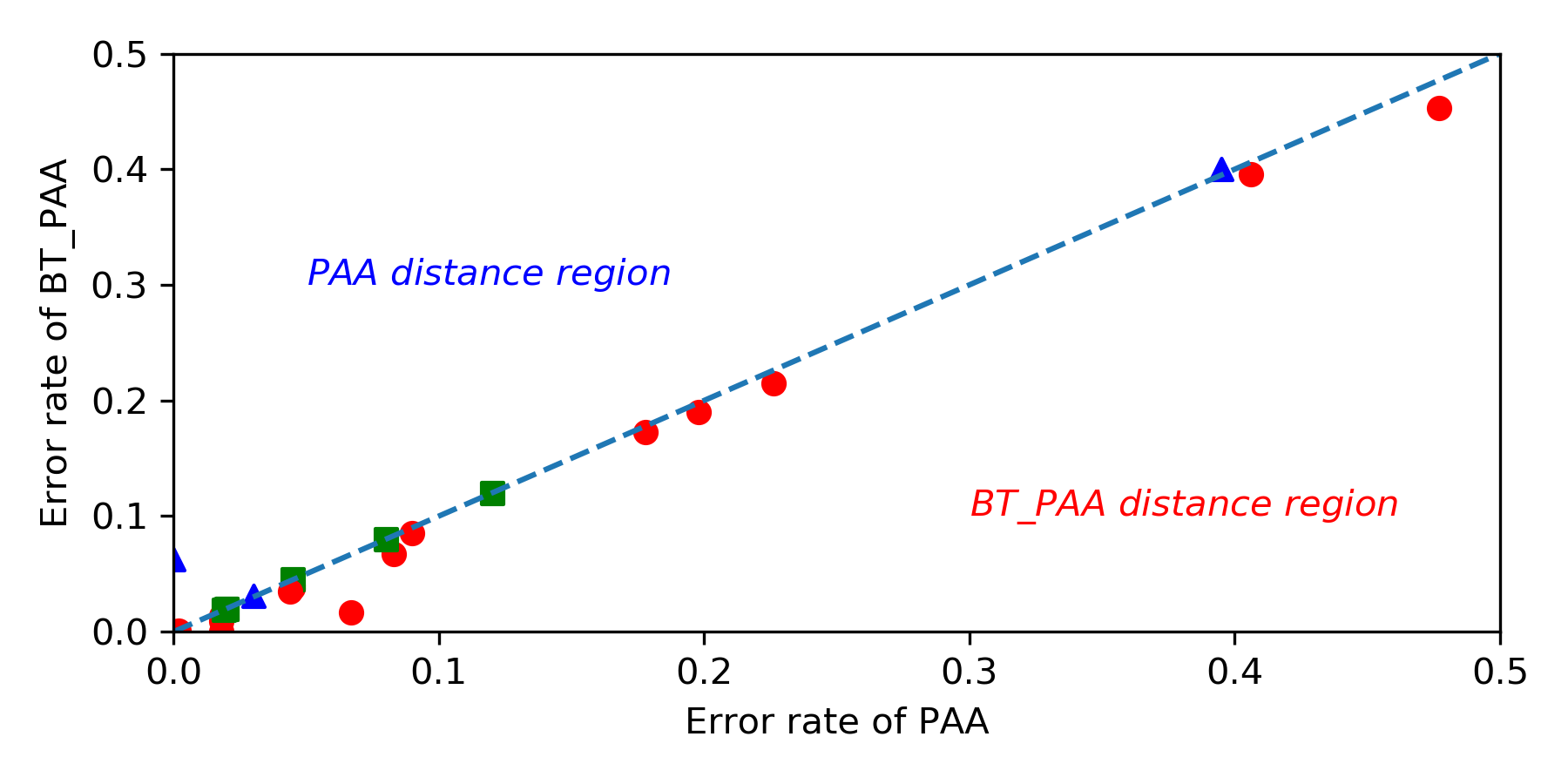} 
	} 
	\caption{ Comparison of error rate between our proposed methods(NT\_PAA and 
	BT\_PAA) and other methods(PAA and ED) with 24 data sets. The red dots in 
	below region represent that our method is superior to the existing one, the 
	blue triangles in up region represent that existing methods are better than 
	ours, and the green squares represent the equal error rate.} 
	\label{fig7} 
\end{figure}

In 
additional, we summarize the result between two methods we mentioned 
above as shown in Fig.\ref{fig7}. If the point(red dot) is in the lower 
region, the proposed methods are more accurate than PAA or ED, otherwise, the 
point(blue triangle) are in up region which means they 
are worsen than original methods. To illustrate the performance, the red dots 
are the majority apparently in four subfigures, so they works well in 
classification via different data sets.

\subsection{Comparison on Anomaly Detection}

In this experiment, we use 12 data sets selected in Table \ref{tab1}, which 
only have two classes, to conduct the anomaly detection experiment with 
algorithm of Local Outlier Factor (LOF)\cite{Breunig2000LOF} to look for 
relatively anomaly points. We use Area Under Curve (AUC) evaluation metric to 
measure the performance. 
The results are shown in Table \ref{tab3} that our proposed measure BT\_PAA 
with 
LOF is much greater than other four distance methods, which five out of twelve 
data sets have a significant increase in AUC, and the other seven have a slight 
increase. As for NT\_PAA, we have eight out of twelve data sets greater than 
other four method. In general, the methods we propose have a much better effect 
than PAA.

\begin{table}[htbp]
	\centering
	\caption{The result of anomaly detection. We choose 12 data sets among UCR 
		which contain only two class and the highest values are highlighted in 
		bold.}
	\begin{tabular}{c|ccccc}
		\toprule
		\multicolumn{1}{c|}{\multirow{2}[4]{*}{data set}} & 
		\multicolumn{5}{c}{AUC} \\
		\cmidrule{2-6}          & \multicolumn{1}{p{4.415em}}{BT\_PAA} & 
		\multicolumn{1}{p{4.415em}}{NT\_PAA} & \multicolumn{1}{c}{PAA} & 
		\multicolumn{1}{c}{Cosin} & \multicolumn{1}{c}{ED} \\
		\midrule
		Coffee & 0.719 & \textbf{0.815} & 0.760 & 0.731 & 0.672 \\
		Computers & 0.582 & \textbf{0.689} & 0.581 & 0.587 & 0.537 \\
		Earthquakes & 0.625 & \textbf{0.673} & 0.590 & 0.665 & 0.585 \\
		ECG200 & \textbf{0.803} & 0.793 & 0.804 & 0.849 & 0.631 \\
		Gun\_Point & 0.639 & \textbf{0.725} & 0.596 & 0.639 & 0.538 \\
		Ham   & \textbf{0.645} & \textbf{0.645} & 0.661 & 0.645 & 0.625 \\
		Herring & 0.623 & \textbf{0.638} & 0.617 & 0.596 & 0.581 \\
		Lighting2 & 0.648 & \textbf{0.653} & 0.646 & 0.657 & 0.579 \\
		ShapeletSim & \textbf{0.960} & 0.863 & 0.961 & 0.712 & 0.900 \\
		Strawberry & 0.606 & \textbf{0.689} & 0.629 & 0.640 & 0.571 \\
		ToeSegmentation2 & \textbf{0.914} & 0.728 & 0.911 & 0.772 & 0.755 \\
		Wine  & \textbf{0.739} & 0.681 & 0.568 & 0.596 & 0.540 \\
		\bottomrule
	\end{tabular}%
	\label{tab3}%
\end{table}%

\begin{figure} \centering 
	\subfigure[] { \label{fig8:b} 
		\includegraphics[width=0.46\textwidth]{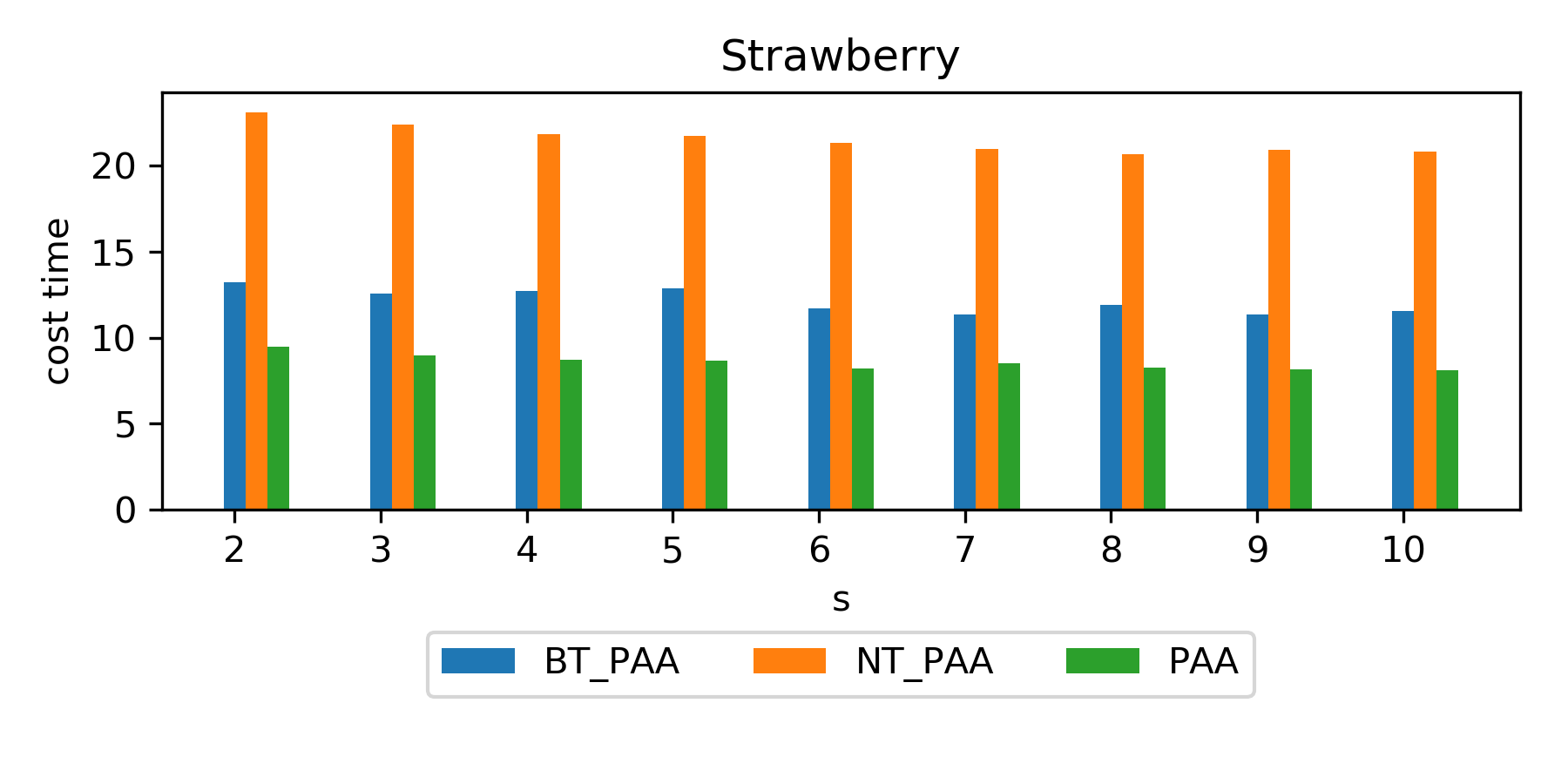} 
	} 
	\subfigure[] { \label{fig8:d} 
		\includegraphics[width=0.46\textwidth]{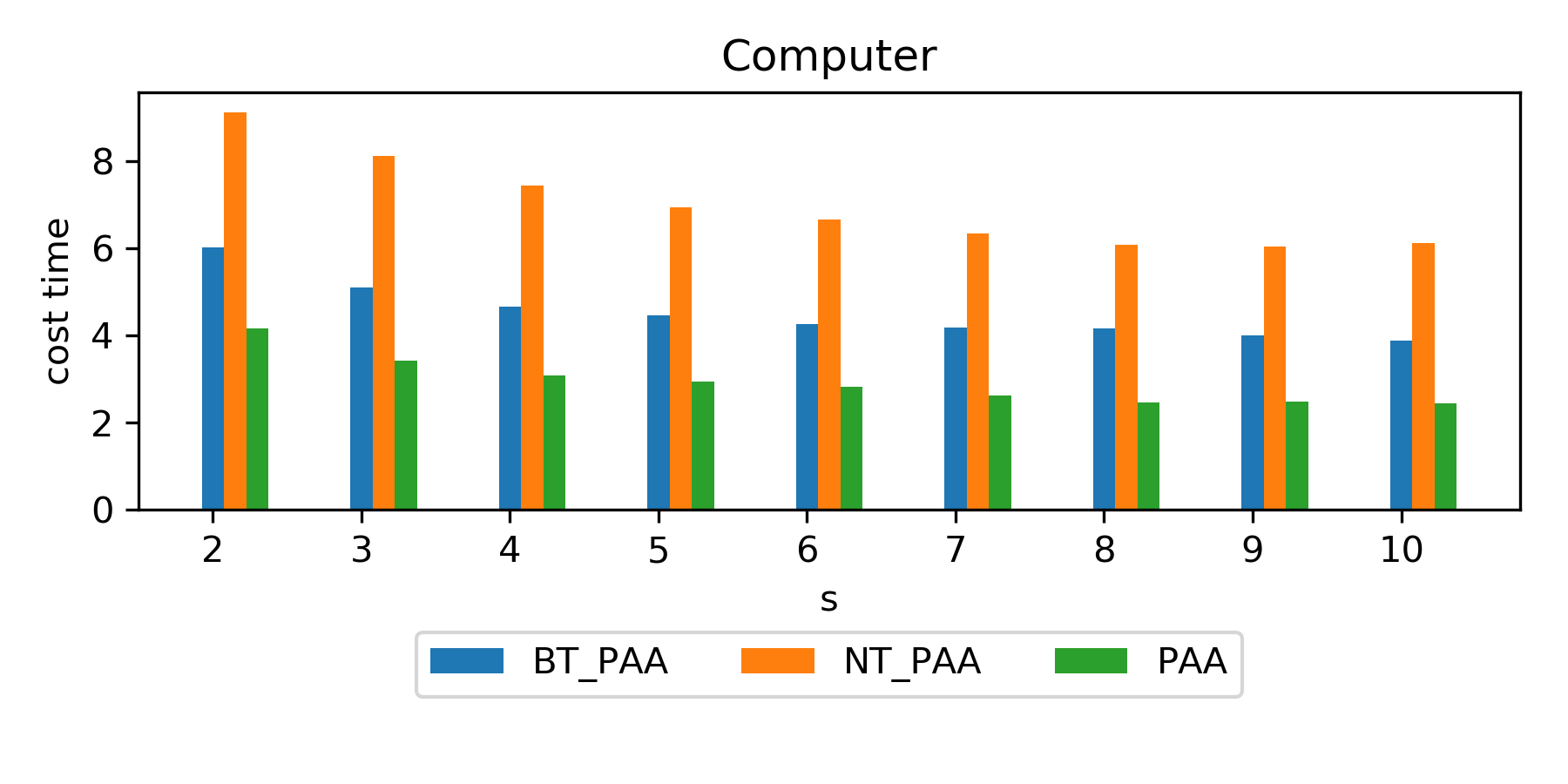} 
	} 
	\caption{ The computation time of different time series with different s 
		ranging from 2 to 10 in anomaly detection.} 
	\label{fig8} 
\end{figure}

To evaluate the computation performance of our two methods, we compare the 
computation time with our methods and PAA in anomaly detection. Five data sets 
are chosen to show the results. From the Fig.\ref{fig8}, the computation time 
of 
NT\_PAA is approximately twice 
as PAA, while the BT\_PAA is a bit lager than PAA, since all three methods have 
same time-consuming in piecewise and the only difference is the time to convert 
time series into binary string and up/below-mean. 
Therefore, BT\_PAA is better 
than NT\_PAA in running time.

\section{Conclusion}
In this paper, we propose two new methods for time series similarity 
measurement that apply trend information.
Our first method use the numerical average in segment which is divided into two 
parts to represent the trend,
and another method use a binary string to record the trend change of a time 
series. And both the trend distance between two sequence are based on the PAA 
distance as the final distance to measure the similarity. We have evaluate the 
proposed methods using the UCR Time Series Archive repository for 
classification 
and anomaly detection, and from the view of the accuracy shows that the 
proposed methods are better than others in both two aspects, despite it costs 
more time than PAA. In our future work, we are planing to reduce the trend 
space and improve the run time of trend distance by using hashing.

\section*{Acknowledgment}
This study is supported by the Shenzhen Research Council (Grant 
No. JSGG2017-0822160842949, JCYJ20170307151518535)

%
%
%
\bibliographystyle{splncs04}

\bibliography{ref}

\end{document}